\documentclass[sigplan,screen]{acmart}
\AtBeginDocument{%
  \providecommand\BibTeX{{%
    \normalfont B\kern-0.5em{\scshape i\kern-0.25em b}\kern-0.8em\TeX}}}

\usepackage{multirow}
\usepackage{subcaption}
\usepackage{colortbl}

\usepackage{xcolor}

\renewcommand{\textit}[1]{\emph{#1}}

\usepackage{fancyvrb}

\copyrightyear{2023}
\acmYear{2023}
\setcopyright{acmlicensed}
\acmConference[FDG 2023]{Foundations of Digital Games 2023}{April 12--14, 2023}{Lisbon, Portugal}
\acmBooktitle{Foundations of Digital Games 2023 (FDG 2023), April 12--14, 2023, Lisbon, Portugal}
\acmPrice{15.00}
\acmDOI{10.1145/3582437.3587211}
\acmISBN{978-1-4503-9855-8/23/04}

\begin{document}
\title{Level Generation Through Large Language Models} 

 \author{Graham Todd}
 \email{gdrtodd@nyu.edu}
 \affiliation{%
   \institution{New York University Tandon}
   \city{Brooklyn}
   \state{New York}
   \country{USA}
 }

 \author{Sam Earle}
 \email{se2161@nyu.edu}
 \affiliation{%
   \institution{New York University Tandon}
   \city{Brooklyn}
   \state{New York}
   \country{USA}
}

 \author{Muhammad Umair Nasir}
 \email{umairnasir1@students.wits.ac.za}
 \affiliation{%
   \institution{University of the Witwatersrand}
   \city{Johannesburg}
   \country{South Africa}
 }

 \author{Michael Cerny Green}
 \email{mike.green@nyu.edu}
 \affiliation{%
   \institution{New York University Tandon}
   \city{Brooklyn}
   \state{New York}
   \country{USA}
 }

 \author{Julian Togelius}
 \email{julian@togelius.com}
 \affiliation{%
   \institution{New York University Tandon}
   \city{Brooklyn}
   \state{New York}
   \country{USA}
 }

%%
%% By default, the full list of authors will be used in the page
%% headers. Often, this list is too long, and will overlap
%% other information printed in the page headers. This command allows
%% the author to define a more concise list
%% of authors' names for this purpose.
\renewcommand{\shortauthors}{Todd, et al.}
% \renewcommand{\shortauthors}{Anonymous}

%%
%% The abstract is a short summary of the work to be presented in the
%% article.
\begin{abstract}
Large Language Models (LLMs) are powerful tools, capable of leveraging their training on natural language to write stories, generate code, and answer questions. But can they generate functional video game levels? Game levels, with their complex functional constraints and spatial relationships in more than one dimension, are very different from the kinds of data an LLM typically sees during training. Datasets of game levels are also hard to come by, potentially taxing the abilities of these data-hungry models. We investigate the use of LLMs to generate levels for the game \textit{Sokoban}, finding that LLMs are indeed capable of doing so, and that their performance scales dramatically with dataset size. We also perform preliminary experiments on controlling LLM level generators and discuss promising areas for future work. 
\end{abstract}

%%
%% The code below is generated by the tool at http://dl.acm.org/ccs.cfm.
%% Please copy and paste the code instead of the example below.
%%

%%
%% Keywords. The author(s) should pick words that accurately describe
%% the work being presented. Separate the keywords with commas.
% \keywords{pcg, sokoban, language models}

\maketitle

\section{Introduction}

In recent years, attention-based large language models (LLMs) have taken the world by storm, demonstrating surprisingly high performance on a variety of natural language tasks. With the right tuning, LLMs have been shown to generate coherent text in a number of styles, produce working snippets of computer code, and even respond naturalistically to human questions and conversation. While the \textit{architectures} underlying these models have been leveraged for tasks outside the realm of standard text generation, from music \cite{huang2018music} to reinforcement learning \cite{chen2021decision}, comparatively less effort has been spent on analyzing the capacity of the LLMs themselves to produce non-lingusitic artifacts while still leveraging their vast amounts of training data. In this paper, we investigate the ability of LLMs to generate video game levels and the extent to which truths about these models taken from natural language processing apply to this new domain. We also conduct preliminary experiments on the capacity to control the levels generated by LLMs using simple data augmentation and prompting.

\begin{figure}
    \centering
    \includegraphics[width=0.5\columnwidth]{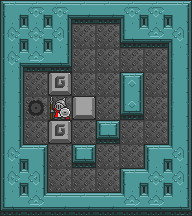}
    \caption{A level for the puzzle game \textit{Sokoban} generated by GPT-3, visualized with the Griddly tileset \cite{bamford2021griddly}}
    \label{fig:my_label}
\end{figure}

Despite their impressive performance, there are reasons to doubt that LLMs would be well suited to the task of level generation. The first is representational. For context, the last few years have seen a steady increase in the use of machine learning to generate novel game content, including game levels. This procedural content generation through machine learning (PCGML) has made use of a variety of methods, including cellular automata, Markov models, convolutional neural networks, and generative adversarial networks. While dissimilar in function, these methods are nonetheless unified in that they tend to represent game levels spatially, as arrangements of tiles or features in two or three dimensions. This is an intuitive approach, as it allows models to more readily learn the local spatial dynamics present in game environments. By contrast, LLMs process inputs and generate outputs in a linear fashion. Game levels must be presented as a sequence of tokens in order to be fed into the model, and generated outputs must be reinterpreted as spatial data in order to be used. Further, variable-length tokenization schemes used by modern LLMs mean that two levels of the same size might be represented with different amounts of underlying tokens. Maintaining regularity and spatial relationships (e.g. attempting to place an enemy directly beneath a player in two dimensions) therefore requires more than simply counting the number of tokens. 
Nevertheless, prior work on n-grams and recurrent neural networks has demonstrated that game levels \textit{can} be represented sequentially without the loss of critical spacial dependencies, albeit with some difficulty. 
We investigate the extent to which this holds for modern, attention-based models and their typical tokenization schemes.

The second potential issue with using LLMs to generate game levels is that of data. 
LLMs are notoriously data hungry, while datasets of game levels are notoriously small, difficult to obtain, and lacking in standardization. The first obvious question is whether the ability of an LLM to generate levels depends on its receiving vast amounts of high-quality training data. More subtly, it is also important to determine whether the vast amount of data used in pretraining actually assists the LLM in producing game levels. It is not clear whether the patterns and structures learned from exposure natural language (or, in some cases, code) transfer to the functional and spatial constraints of game levels. The quality and even playability of a game level is often dependent on factors such as topology or the relative amounts of different tile types -- a far cry from the syntax of English or Python! 

Even so, LLMs also seem to have certain advantages when it comes to level generation, namely controllability and generalizability. Controllability here refers to the possibility of using natural language prompts to generate levels with particular characteristics. Recent work has demonstrated that natural language-guided generation is possible not only for text, but also for images \cite{ramesh2022hierarchical} and music \cite{agostinelli2023musiclm}. These systems leverage LLMs and are capable of accommodating a wide range of potential prompts, an impressive feat that provides some reason for optimism that current approaches for controllable level generation could be similarly improved. At the same time, LLMs have shown considerable promise in generalizing to unseen domains \cite{gpt3} or across a wide variety of tasks \cite{reed2022generalist}. With respect to level generation, this might allow for a single LLM-based model to produce levels for multiple games or even, with sufficiently detailed prompting, a previously unencountered game. 

In this paper, we aim to answer some of the initial questions surrounding the ability of LLMs to generate game levels using the iconic puzzle game \textit{Sokoban}. We perform experiments on the effects of pretraining and dataset size, as well as a preliminary investigation on the controllability of LLM level generators. We conclude with a discussion of the results and the many fruitful avenues for future work.

\begin{figure}
\begin{subfigure}[t]{.47\columnwidth}
    \centering
    \includegraphics[width=\linewidth]{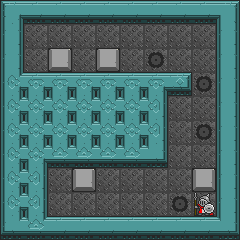}
    \caption{Playable level generated by GPT-2}
    \label{fig:my_label}
\end{subfigure}
\begin{subfigure}[t]{.47\columnwidth}
    \centering
    \includegraphics[width=\linewidth]{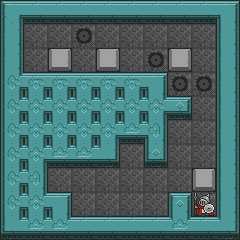}
    \caption{Nearest level (edit distance) in the \texttt{Boxoban} set}
    \label{fig:my_label}
\end{subfigure}
\caption{A novel and playable generated level, and its nearest neighbor in the training set.}
\end{figure}

\section{Related Work}

Procedural content generation (PCG), refers to the use of automated or algorithmic methods to create artifacts, typically for use in art or games. Techniques for PCG range from simple noise functions to complex neural models. Our work falls into the broad category of procedural content generation via machine learning (PCGML) \cite{summerville2018procedural}, in which content generating functions are learned from extant datasets. Liu et al. present an overview of the PCGML field, with a specific focus on deep learning \cite{liu2021deep}. 

For the specific task of generating game levels, common model choices include variational autoencoders \cite{snodgrass2020multi}, generative adversarial networks \cite{park2019generating,torrado2020bootstrapping}, evolution \cite{schrum2020interactive}, and reinforcement learning \cite{khalifa2020pcgrl}. In addition, however, there is a history of using autoregressive models typically found in natural language processing for game level generation. Dahlskog et al. present early work on this approach, using a simple n-gram approach to generate novel Super Mario Bros. levels from an existing dataset by treating a level as a left-to-right sequence of ``tokens'' each representing a vertical slice \cite{dahlskog2014linear}. This work was quickly expanded to use long short-term memory (LSTM) networks \cite{summerville2016lstm}, a model choice which has also found success in generating levels based on human play traces \cite{summerville2016learning} and combining levels from multiple games \cite{sarkar2018blending}.

Our work also borrows from the literature on controllable PCG, in which specific parameters are provided to the generator in order to guide its outputs. Approaches for controllable PCG often involve manipulating a latent embedding vector, with prior work having made use of generative networks \cite{mott2019controllable}, VAEs \cite{sarkar2020conditional, sarkar2021dungeon}, and reinforcement learning \cite{earle2021learning}. 

Our target domain of \textit{Sokoban} is a popular choice for PCG work, ranging from early rule- and template-based approaches~\cite{sokobantemplates,sokobanrules}, to search-based methods \cite{sokobanmcts} and recurrent neural networks \cite{sokobanrnn}. Zakaria et al. present a particularly in-depth comparison of various PCG methods for \textit{Sokoban}, including LSTMs \cite{zakaria2022sokoban}. They bootstrap an initially-small dataset of levels and use it to train their level generators, demonstrating that LSTMs are capable of producing a variety of novel and playable levels. They also perform experiments on the controllability of their generators. Their results indicate that while the task is challenging, LSTMs, in addition to other existing PCGML methods, can adhere to specified characteristics at levels substantially above chance. Our analysis is similar, though we instead focus our attention on a more modern class of large language model. In addition to a change in architecture, we also specifically interrogate some of the standard assumptions about the training and behavior of LLMs, and whether or not they are helpful for the task of level generation.

Finally, contemporaneous work makes use of a transformer-based language model to controllably generate levels for Super Mario Bros \cite{sudhakaran2023mariogpt}. This investigation is complementary to ours, and focuses on a different target domain.

\section{Data}

We train our models on levels from the game \textit{Sokoban}, a block-pushing puzzle game released in 1982 by Thinking Rabbit. In \textit{Sokoban}, the player is tasked with navigating along a rectangular grid in order to push boxes into specified target squares. A level can easily be represented as a grid of ASCII characters, where each character is mapped to either: a wall, an empty space, the player, a box, a goal, a box on top of a goal, or a player on top of a goal. Despite its simplicity of representation, \textit{Sokoban} levels can be very challenging for both human and artificial agents alike, owing to its rapidly-branching state space and the fact that certain game states are ``unrecoverable'' and, once reached, cannot be escaped from.

We use two sets of \textit{Sokoban} levels to train our models. The first is the \texttt{Microban}\footnote{Microban dataset available here: \href{https://tinyurl.com/yckwxd7k}{https://tinyurl.com/yckwxd7k}} dataset, consisting of roughly 500 levels created by David W. Skinner.  We restrict our dataset to 282 levels for which an ASTAR search agent was able to find a solution. Levels in this set range in size from 5 by 3 to 27 by 12, with solution lengths ranging from 1 to 279.

The second set of \textit{Sokoban} levels is the \texttt{Boxoban} dataset, which consists of 438,000 procedurally generated levels. Levels were generated using a combination of heuristic and pattern-based rules \cite{boxoban}. Unlike with the \texttt{Microban} set, levels in the \texttt{Boxoban} set are all 10 by 10 and contain 4 boxes / goals. We use the ASTAR agent to recover solutions for all 438,000 levels, and solution lengths range from 6 to 206.

To construct our dataset, we split each level into a set of lines (applying a padding of walls in the case of non-rectangular levels), concatenate them to form a single string of characters, and finally apply the model's tokenizer.

\section{Models}
\label{model}

Our core experimental model is the Generative Pre-Trained Transformer (GPT), a class of attention-based language model \cite{gpt2,gpt3}. Both GPT-2 and GPT-3 are trained by attempting to predict the next ``token'' (typically a word or word piece) given the context of preceding tokens. Owing to its greater availability, we focus the majority of our experiments on GPT-2 and variants thereof.

\section{Metrics}
\label{metrics}

To measure the ability of LLMs to generate game levels, we use the following four metrics:
\begin{itemize}
    \item \textit{Playability}: we measure the proportion of generated game levels that are ``valid''. In the case of Sokoban, this means that they are rectangular, contain only valid characters, contain an equal and non-zero number of boxes and goals, contain exactly one player, and are solvable. We determine solvability using an ASTAR tree-search agent. If, after running for 150,000 steps, the ASTAR agent fails to find a valid solution, we deem the level unplayable. This provides a lower bound on the true rate of playability. 
    \item \textit{Novelty}: we measure the proportion of generated game levels that are distinct from each level in the training dataset. We use a simplified approach that treats two levels as distinct if their string edit distance is above some threshold. We note, however, that this definition of novelty does not take into account \textit{functional} differences between levels (e.g. two levels may differ in only a single tile but nonetheless have very different solutions). Exploring the effects of different novelty measures remains an interesting area for future work. For our experiments, we use a the edit-distance approach and a threshold of $k = 5$.
    \item \textit{Diversity}: we measure the proportion of generated game levels that are mutually distinct from each other. Using the same definition of distinctness as above, we use a graph-based approach to find the largest subset of generated levels that are all at least $k = 5$ edit distance from each other. Specifically, we convert the set of generated levels into an undirected graph where two nodes (levels) have an edge if their edit distance is above the threshold $k$. We then find the largest clique (subset of fully connected nodes) on this graph, and report the diversity as the size of this clique divided by the number of generated levels (a set of levels on this graph is only fully connected if each level is at least $k$ edit distance away from every other level in the set). Because finding a maximum clique can be computationally intractable, we terminate the clique-finding algorithm after a specified number of iterations (1 million) and report the size of the largest clique found. This provides a lower bound on the model's diversity. 
    \item \textit{Accuracy}: in the case of controllability experiments, we measure the proportion of generated game levels that adhere to the given prompt. Rather than enforce an exact match between prompt and output, we allow the model to generate levels that are within a certain experiment-specific tolerance of the specified characteristic (for instance, with a tolerance of 5 we would consider accurate a level with a solution length of 21, when the prompt called for a level with solution length 25)
\end{itemize}

\section{Experiments}
\subsection{Effects of Pretraining}
\label{pretraining_exp}

Our first experiment aims to answer two questions:
\begin{enumerate}
    \item Are LLMs capable of generating novel, playable game levels?
    \item Does the extensive pretraining given to LLMs affect their ability to generate game levels?
\end{enumerate}

LLMs are typically trained on vast quantities of text collected from a variety of natural language contexts and then later ``fine-tuned" on a smaller, more task-specific dataset. 
While pretraining has been shown to improve models' performance on a variety of downstream linguistic tasks, it is less clear whether it would help in the more specialized task of generating valid game levels.
To examine this question, we consider 3 variants of the GPT-2 model: \textit{standard}, \textit{java-adapted}, and \textit{untrained}. The \textit{standard} GPT-2 model was pretrained on the \texttt{WebText} dataset, consisting of the content of 45 million links \cite{gpt2}, the \textit{java-adapted} model was pretrained on the \texttt{CodeSearchNet} dataset of Java code, consisting of 1.6 million Java methods \cite{codegpt}, and the \textit{untrained} model, unsurprisingly, received no pretraining (weights are randomly initialized). As a note, both \textit{standard} and \textit{java-adapted} models use specialized tokenizers which are trained to efficiently break input strings into sub-word tokens. Rather than use an existing tokenizer, we allowed the \textit{untrained} model to train a custom tokenizer on the game level dataset, using the same byte-pair encoding scheme as GPT-2. 

Each model is trained for 100k steps with 5 random seeds on the \texttt{Boxoban} dataset with the following training hyperparameters: learning rate of 0.0001, weight decay of 0.0001, a batch size of 32, and the \texttt{AdamW} optimizer.Each model took roughly 24 hours to train on a single A100 GPU. In order to evaluate a model, we provide it with some initial context (for this experiment, only the \texttt{START} token) and then use beam search with random sampling in order to generate one or more continuations. We then compute the proportion of generated levels that are novel, the proportion that are playable, and the proportion that are novel, playable, and diverse. For simplicity, we call the proportion of novel, playable, and diverse levels the model's \textbf{score} (e.g. if the model produces 54 levels that are playable and novel out of 100 samples, of which 47 are mutually distinct, we report a score 0.47).

Because the outputs of a LLM are largely dependent on the hyperparameters used during generation, for each model and seed we perform an additional sweep over the generation \textit{temperature}, the \textit{top-p} value, and the number of \textit{beams}. Each inference takes only a few minutes on a single A100 GPU. The entire sweep was completed in roughly 2 hours. For each model, we select the evaluation hyperparameters which achieve the highest score when averaged over the 5 random seeds. We report these average scores, along with the average novelty, playability, and diversity rates, for each model in Table~\ref{tab:exp_1_results}. 

\subsection{Effects of Dataset Size}
\label{dataset_size_exp}

Another well-known property of LLMs is that their performance on a variety of NLP tasks tends to scale with the amount of training data \cite{kaplan2020scaling}, but does this trend hold for the specialized task of generating game levels? This question is particularly important because in many situations it is difficult or impossible to collect a large set of high-quality game levels. Relatedly, in situations where large amounts of game levels \textit{are} available (typically games for which heuristic or rule-based PCG approaches exist), do LLMs benefit from ever-increasing dataset sizes? Finally, can simple data augmentation approaches improve LLM performance?

First, we consider four ``slices'' of the \texttt{Boxoban} dataset consisting of 0.1\%, 1\%, 10\%, 100\% (i.e. the complete dataset used above) of the data, randomly sampled. We take the standard GPT-2 model and re-train it on each of the slices for 100k steps, using the same training hyperparameters as above. We then evaluate each model's novelty, playability, and ``score'' using the same procedure as in Section~\ref{pretraining_exp}. As before, we find the evaluation hyperparameters that achieve the highest average score for each model, and report the results in Table~\ref{tab:dataset_size}.

Next, we train a GPT-2 model on the \texttt{Microban} dataset, as well as two augmented versions of the dataset: $\texttt{Microban}_{\texttt{flip}}$ (levels flipped about the X and Y axes) and $\texttt{Microban}_{\texttt{flip+rotate}}$ (levels rotated 90 degrees clockwise and counterclockwise). Each model is trained for 100k steps, and the highest achieved average score for each is reported in Table~\ref{tab:microban}.

\begin{table}
\begin{center}
\resizebox{\columnwidth}{!}{
\begin{tabular}{lrrrr}
\toprule
 & Novelty & Playability & Diversity & Score \\
Model &  &  &  &  \\
\midrule
\textit{GPT-2} & \bfseries 0.97 & 0.54 & \bfseries 1.00 & 0.53 \\
\textit{GPT-2 (Untrained)} & 0.96 & \bfseries 0.60 & \bfseries 1.00 & \bfseries 0.56 \\
\textit{Java GPT-2} & \bfseries 0.97 & 0.54 & \bfseries 1.00 & 0.53 \\

\bottomrule
\end{tabular}
}
\end{center}
\caption{Novelty, playability, diversity, and overall ``score'' (defined as the diversity of the subset of generated levels that are both novel and playable) for each type of pretraining, using the best evaluation hyperparameters when averaged over 5 seeds.}

\label{tab:exp_1_results}
\end{table}

\subsection{Controllability}
\label{controllability_exp}

Arguably the most compelling reason to use LLMs for game level generation is the ability to prompt the model in natural language to generate levels with specific characteristics. For instance, it might be possible to create a level that has a specific difficulty (represented by the length of the solution), or with certain level topologies. Recent LLMs have demonstrated impressive abilities to leverage prompting in order generalize from few or even zero examples on a variety of tasks \cite{gpt3}. However, zero-shot generalization is likely to be difficult for level generation owing to the many functional constraints on valid game levels and their dissimilarity from inputs encountered during pretraining. Thus, we instead focus on LLMs that have been trained specifically to adhere to prompts during level generation. We accomplish this by simply prepending an ``annotation'' to each level in the training dataset. Two examples of annotated levels are presented in Figure~\ref{fig:level_example}. At generation time, we provide the model with only the annotation and task it with generating the rest of the level while adhering to the specified values.

\begin{figure}
    \begin{subfigure}[b]{0.23\textwidth}
        \begin{center}
        \verb|prop_empty: 0.25|
        
        \verb|solution_len: 65|
        
        \verb|########|
        
        \verb|##----##|
        
        \verb|##.-..##|
        
        \verb|###$-@-#|
        
        \verb|#-$--$-#|
        
        \verb|#---####|
        
        \verb|########|
        \end{center}
    \end{subfigure}
    \hfill
    \begin{subfigure}[b]{0.23\textwidth}
        \begin{center}
        \verb|prop_empty: 0.269|
        
        \verb|solution_len: 42|
        
        \verb|#########|
        
        \verb|#---#####|
        
        \verb|#---#---#|
        
        \verb|##-$*@--#|
        
        \verb|##-*.--##|
        
        \verb|##--#####|
        
        \verb|#########|
        \end{center}

    \end{subfigure}
    \caption{Two levels taken from the \texttt{Microban} dataset, along with their annotations}
    \label{fig:level_example}
\end{figure}

For this experiment, we focus on two annotated characteristics: the \textit{proportion of empty space} (i.e. percentage of level tiles that are not players, walls, boxes, or targets) and the \textit{solution length}. Both of these are measurable characteristics of valid \textit{Sokoban} levels, though they differ in complexity. The proportion of empty space is an \textit{observable} characteristic of a level, requiring only the ability to count in order to compute. Solution length, by contrast, can typically only be computed by actually solving the level in question and not through direct observation. Even visually sparse or simple levels can require long solutions.

As with the dataset size experiment in Section~\ref{dataset_size_exp}, we use a standard pretrained GPT-2 model. We train a separate model on the \texttt{Boxoban} dataset annotated with the proportion of empty space and the \texttt{Boxoban} dataset annotated with level solution length. At test time, we provide the model with only the annotation, randomly sampled from the collection of annotations in the training set. In addition to novelty, playability, and diversity, we compute the model's accuracy as described in Section~\ref{metrics}. For the proportion of empty space condition, we use a tolerance of 0.01, and for the solution length condition we use a tolerance of 5. For this experiment, we report both the standard ``score'' defined above, as well as the ``control score,'' which is simply the diversity of levels that are accurate to the prompt, in addition to being novel and playable. We report these results, using the same evaluation procedure as in Section~\ref{pretraining_exp}, in Table~\ref{tab:controllability}.

\begin{table}
\begin{center}
\resizebox{\columnwidth}{!}{
\begin{tabular}{lrrrr}
\toprule
 & Novelty & Playability & Diversity & Score \\
\% of \texttt{Boxoban} &  &  &  &  \\
\midrule
0.1\% & 0.00 & \bfseries 0.80 & 0.01 & 0.01 \\
1\% & 0.10 & 0.66 & 0.97 & 0.03 \\
10\% & 0.90 & 0.55 & \bfseries 1.00 &  0.47 \\
100\% & \bfseries 0.97 & 0.54 & \bfseries 1.00 & \bfseries 0.53 \\
\bottomrule
\end{tabular}
}
\end{center}
\caption{Novelty, playability, diveristy, and overall score for GPT-2 trained on increasing amounts of the \texttt{Boxoban} dataset, using the best hyperparameters averaged over 5 seeds. Increasing dataset size leads to increased performance.}
\label{tab:dataset_size}
\end{table}

\subsection{Preliminary Investigation on GPT-3}

While GPT-2 has demonstrated very high performance on a variety of natural language tasks, it has nonetheless been largely eclipsed by its successor: GPT-3, which boasts both substantially more parameters as well as a greatly increased amount of pretraining data. Access to GPT-3 is currently limited, making it infeasible to perform direct comparisons with GPT-2 on all measures. Nevertheless, we perform some initial experiments on the performance of OpenAI's \textit{Davinci} model when trained on the \texttt{Microban} dataset and its augmentations.

We train the \textit{Davinci} model for 10 epochs separately on each of the datasets using a single seed. At test time, we perform a limited hyperparameter sweep over generation temperature and top-p. As with GPT-2, we compute the model's novelty, playability, and overall score. We report the GPT-3 results in Table~\ref{results:gpt3}.

\begin{table}
\begin{center}
\resizebox{\columnwidth}{!}{%
\begin{tabular}{lrrrr}
\toprule
 & Novelty & Playability & Diversity & Score \\
Dataset &  &  &  &  \\
\midrule
\texttt{Microban} & \bfseries 0.59 & 0.30 & 0.83 & 0.02 \\
$\texttt{Microban}_{\texttt{flip}}$ & 0.56 & 0.32 & \bfseries 0.89 & 0.02 \\
$\texttt{Microban}_{\texttt{flip+rotate}}$ & 0.24 & \bfseries 0.54 & 0.82 & \bfseries 0.04 \\
\bottomrule
\end{tabular}
}
\end{center}
\caption{Novelty, playability, diversity, and overall score for GPT-2 trained on the \texttt{Microban} dataset and two augmentations, using the best hyperaparmeters averaged over 5 seeds. The model broadly overfits and fails to generate novel and playable levels.}

\label{tab:microban}
\end{table}
% \vspace{-1cm}

\section{Results}
\subsection{Effects of Pretraining}

We see in Table~\ref{tab:exp_1_results} that all three models are able to generate novel, playable, and diverse levels. An average ``score'' of around 0.55 indicates that the language model is able to reliably generate \textit{Sokoban} levels that are valid and solvable without directly copying from its \texttt{Boxoban} training dataset. We observe that the untrained GPT-2 model performs very slightly better than either of the pretrained models. The difference, however, is minute and likely to the effect of random variance. Overall, this seems to indicate that the pretraining afforded to these LLMs neither particularly helps nor hinders their ability to generate game levels. This could be explained by the substantial dissimilarity between modeling natural language and \textit{Sokoban} levels, meaning that models are required to effectively learn from scratch in this domain and are able to do so.

\begin{table*}
\begin{center}
\begin{tabular}{lrrrrrr}
\toprule
 & Novelty & Playability & Accuracy & Diversity & Score & Control Score \\
Controls &  &  &  &  &  &  \\
\midrule
Prop. Empty & \bfseries 0.96 & 0.57 & \bfseries 1.00 & 0.97 & \bfseries 0.53 & \bfseries 0.53 \\
Solution Len & 0.95 & 0.54 & 0.17 & \bfseries 1.00 & 0.50 & 0.14 \\
Prop. Empty \& Solution Len & \bfseries 0.96 & \bfseries 0.59 & 0.03 & 0.79& 0.45 & 0.03 \\
\bottomrule
\end{tabular}
\end{center}
\caption{Novelty, playability, diversity, and accuracy (along with the overall score and the ``control score'', which accounts for accuracy) for GPT-2 trained on \texttt{Boxoban}, annotated with the proportion of empty space, the solution length, and both simultaneously, using the best hyperparameters when averaged over 5 seeds. The model is able to adhere to the empty space controls, but not the solution length controls.}
\label{tab:controllability}
\end{table*}

\subsection{Effects of Dataset Size}
\label{results:dataset_size}

The results in Table~\ref{tab:dataset_size} and Table~\ref{tab:microban} indicate that dataset size is indeed an important factor for an LLM's ability to generate game levels. For small datasets (i.e. the 0.1\% and 1\% conditions of \texttt{Boxoban}, as well as also \texttt{Microban} conditions), GPT-2 can produce levels that are independently novel or playable in isolation, but not levels that are both, leading to low overall scores. Nevertheless, in all but the 0.1\% \texttt{Boxoban} condition, sample diversity remains relatively high. While the effect is not especially pronounced, there does appear to be a correlation between the size of the dataset and the model's score. This supports the notion that, like with many natural language tasks, LLM performance on level generation scales effectively with the availability of training data. 
We note, however, that prior works demonstrates LSTMs are capable of generating novel levels when trained on a bootstrapped dataset consisting originally of only 12 samples \cite{zakaria2022sokoban}, meaning that it is unlikely that LLMs are fully \textit{incapable} of performing well when restrcted to small datasets.
What might account for this difference in performance, then? One possibility is expressivity: modern transformers are much better able to represent sequential data than LSTMs, and so are more likely to completely model the dynamics of their training datasets, to the detriment of their generative capabilities. However, as we will see, this explanation does not account for the performance of GPT-3 (see Section~\ref{gpt3-discussion}).

\subsection{Controllability}

In Table~\ref{tab:controllability}, we see effects of sampling levels conditioned on simple prompts. In the first row, we see that the GPT-2 model is able to produce levels that are novel, playable, and within a single tile of the specified proportion of empty space (corresponding to perfect accuracy and a relatively high control score). However, when it comes to solution length, GPT-2 achieves an accuracy of only ~17\%. Given the tolerance of 5 and the fact that most solution lengths in the dataset fall within a relatively narrow band, this cannot be interpreted as anything more than the effects random chance. A similar fact holds for the combined condition, where overall accuracy is determined by both the correct amount of empty space and solution length and does not rise above 3\%. It is worth noting, however, than even in the conditions where GPT-2 failed to produce accurate levels, it nonetheless continued to generally produce novel and playable ones. In other words, the introduction of the prompt did not negatively affect the model's performance.

\subsection{Preliminary Investigation on GPT-3}
\label{gpt3-discussion}

\begin{table}
\label{tab:microban_gpt3}
\centering
\resizebox{\columnwidth}{!}{%
\begin{tabular}{lrrrr}
\toprule
 & Novelty & Playability & Diversity & Score \\
Dataset &  &  &  &  \\
\midrule
\texttt{Microban} & 0.09 & 0.88 & 0.67 & 0.01 \\
$\texttt{Microban}_{\texttt{flip}}$ & 0.55 & \textbf{0.94} & 0.77 & 0.36 \\
$\texttt{Microban}_{\texttt{flip+rotate}}$ & \textbf{0.70} & 0.93 & \textbf{0.88} & \textbf{0.51} \\
\bottomrule
\end{tabular}
}
\caption{Novelty, playability, diversity, and overall score for GPT-3 trained on the \texttt{Microban} dataset and two augmentations. GPT-3 is able to produce novel, playable, and diverse levels from a relatively small training set.}
\label{results:gpt3}
\end{table}

Table~\ref{results:gpt3} contains the results of GPT-3 level generation when trained on the \texttt{Microban} and its augmentations. While these results should be taken with a healthy amount of caution because they are generated from only a single training run and with a limited evaluation hyperparameter sweep, they nonetheless offer some reason for optimism. In contrast to GPT-2, GPT-3 is able to produce novel and playable levels when trained on both the augmented forms of the \texttt{Microban} dataset, with its overall score on the final condition approaching that of GPT-2 trained on the entire \texttt{Boxoban} dataset. As with previous experiments, however, we observe that increasing dataset size (in this case adding rotations in addition to flips) does lead to increased overall performance with GPT-3. In future work, we intend to perform a more robust analysis of GPT-3's abilities, including its capacity for controllable level generation.

\section{Future Work}

In this paper, we examine the performance of LLMs on generating levels for a single game. However, one of the primary strengths of LLMs is their ability to rapidly adapt to a variety of contexts given the appropriate prompt. Consider a dataset of levels from many different games, where each level has been annotated with the natural language mapping from tiles to game objects (e.g. ``\texttt{@} represents the player, \texttt{M} represents a monster), along with a description of the level objective. An LLM might be better equipped than other PCG systems to generate novel and playable levels from this variety of games, owing to its familiarity with natural language and capacity for rapid adaptation. 

However, our work also indicates that making effective use of LLMs for game level generation may require more consideration of dataset size: few games have available the massive amount of levels present in the \texttt{Boxoban} set. As mentioned in Section~\ref{results:dataset_size}, prior work has demonstrated that bootstrapping larger training sets from initially small collections of levels is a viable technique. Another possibility is augmenting existing datasets beyond simple flips and rotations. More generally, we should consider ``fundamental tension of PCGML'' \cite{karth2019addressing}: at what point does the cost of obtaining training data for automated content generators exceed the cost of making the content by hand? While it's possible that LLMs require too much data to be feasible game content generators, the reasonable performance of GPT-3 on the small \texttt{Microban} dataset offers some optimism that this tension might be ameliorated by more sophisticated models.

It is also important to note that the large amounts of data used to pretrain LLMs could potentially include \textit{Sokoban} levels in various formats. This fact complicates the notion of ``novelty,'' as it's possible for the model to produce levels that are distinct from its fine-tuning dataset but are nonetheless copies of extant game levels. One potential approach for mitigating this danger would be to separate out the prompt encoding and level generation systems and use a model without pretraining for the latter (while still retaining the benefit of pretraining for the component of the model that understands natural language prompts).

Finally, there is room for much greater sophistication in the techniques used to control LLM outputs. Research in the area of controllable language model decoding \cite{li2017learning} offers the opportunity to leverage existing work in PCG through reinforcement learning. More modern LLMs, especially, have also been shown to benefit from careful prompt engineering \cite{zhao2021calibrate}. A combination of these approaches might allow for LLM generators that are better equipped to obey the functional constraints of game levels.

\section{Conclusion}

Large languages models are highly versatile. Beyond merely predicting likely continuations of text, they are capable of an impressive range of natural language tasks. In this work, we show that generating video game levels can be added to that list. With sufficient data and training, LLMs are able to produce a diverse set of novel and playable \textit{Sokoban} levels. We show that the pretraining generally afforded to these models does not hinder its ability to generate game levels, though any actual effect is unclear. We also demonstrate that, for GPT-2, the domain of game level generation is beholden to the same data scaling trends that apply to many natural language domains -- model performance is strongly dependent on the availability of data. Cutting-edge LLMs like GPT-3 may have the potential to better generalize from small amounts of training data, though more work must be done before decisive conclusions can be drawn. With respect to controllability, we find that a simple prompting approach is sufficient for observable level characteristics like the proportion of empty tiles, but breaks down on more complicated metrics like solution length. Overall, the use of LLMs for game level generation shows promise despite it being a wildly different domain from natural language, complete with its own set of constraints and syntax. LLMs also seem potentially poised to overcome the general lack of available game level data, potentially offering a new way forward for procedural content generation through machine learning.

\appendix

\section*{Ethical Statement}
Large language models have known biases and limitations, and can occasionally produce harmful or toxic text. While our models are trained to produce game levels, such training does not entirely eliminate this possibility. In addition, we note the possibility of LLMs copying published game levels included in their pretraining corpuses, a fact which should be considered before any form of widespread or commercial adoption.

\bibliographystyle{ACM-Reference-Format}
\bibliography{bibliography}

\end{document}